\documentclass[runningheads]{llncs}
\usepackage[T1]{fontenc}
\usepackage{graphicx}
\usepackage{graphicx}
\usepackage{amsmath}
\usepackage{amssymb}  
\usepackage{booktabs}
\usepackage{multicol}
\usepackage{multirow}
\usepackage{threeparttable}
\usepackage{bigdelim}
\usepackage{xcolor}
\usepackage{float}
\usepackage{makecell}
\usepackage{hyperref}

\newcommand{\tblresultCombo}{
\begin{table}
  \caption{Classification accuracy of our NNS action recognition model, under various convolutional and temporal configurations and two image modalities. We test on the NNS clinical in-crib data under subject-wise leave-one-out cross-validation, and on the NNS in-the-wild data directly, both with balanced classes. Strongest results in bold.}
    \centering
    \ra{1.0} 
    \resizebox{\columnwidth}{!}{\begin{tabular}{@{}llrrrrrrrrrrr@{}}
        \toprule
        & \phantom{a} & \phantom{a} & \multicolumn{4}{c}{\text{Optical Flow}} & \phantom{a} & \multicolumn{4}{c}{\text{RGB}}\\ 
        \cmidrule[0.5pt]{4-7} \cmidrule[0.5pt]{9-12}
        & & Convolutional & 1-lr. CNN & ResNet18 & ResNet50 & ResNet101 && 1-lr. CNN & ResNet18 & ResNet50 & ResNet101\\
        Dataset & Sequential & \# Tr. Params. & 333K & 154K & 614K & 614K && 333K & 154K & 614K & 614K \\
        \midrule
        \multirow{3}*{\makecell[l]{Clinical}} 
          & Transformer & 393K & 90.9 & 89.4 & 88.5 & 89.2 &                    & \textbf{63.5} & 53.5 & 56.4 & 47.3\\
          ~ & LSTM        &  418K & 90.7 & \textbf{94.9} & 87.9 & 85.2 &  & 52.9 & 52.1 & 57.5 & 46.8\\
		~ & Bi-LSTM     & 535K  & 86.5 & 94.5 & 90.6 & 91.4 &           & 56.2 & 46.3 & 53.5 & 50.4\\

        \midrule
        \multirow{3}*{\makecell[l]{In-the-\\wild}} & Transformer & 393K  & 83.6 & 79.5 & 81.4 & \textbf{92.3} &  & 54.0 & 53.3 & 48.9 & \textbf{59.4}\\
        ~  & LSTM        & 418K   & 84.5 & 80.8 & 84.6 & 82.7 &                    & 50.5 & 55.0 & 50.2 & 50.2\\
		~ & Bi-LSTM     & 535K  & 87.2 & 85.2 & 87.5 & 87.2 &           & 54.4 & 51.7 & 50.2 & 49.8\\
        
        \bottomrule
    \end{tabular}}
    \label{tab:resultCombo}
\end{table}

\newcommand{\tblsegNew}{
\begin{table}
  \caption{Average precision $\text{AP}_t$ and average recall $\text{AR}_t$ performance for various IoU thresholds $t$ of our NNS segmentation model. We test three local classification aggregation methods and two different classifier confidence thresholds. Precision-recall pairs with the highest precision in each threshold configuration in bold.}
    \centering
    \ra{1.0} 
    \resizebox{\columnwidth}{!}{\begin{tabular}{@{}llrrrrrrrrrrrrrrr@{}}
        \toprule
        & \phantom{a} & & \multicolumn{6}{c}{\text{Classifier Confidence Threshold = 0.9}} & \phantom{a} & \multicolumn{6}{c}{\text{Classifier Confidence Threshold = 0.5}}\\ 
        \cmidrule[0.5pt]{4-9} \cmidrule[0.5pt]{11-16}
        Dataset & Method && $\text{AP}_{0.1}$ & $\text{AR}_{0.1}$ & $\text{AP}_{0.3}$ & $\text{AR}_{0.3}$ & $\text{AP}_{0.5}$ & $\text{AR}_{0.5}$ & & $\text{AP}_{0.1}$ & $\text{AR}_{0.1}$ & $\text{AP}_{0.3}$ & $\text{AR}_{0.3}$ & $\text{AP}_{0.5}$ & $\text{AR}_{0.5}$ \\
        
        \midrule
        \multirow{3}*{\makecell[l]{Clinical}}  & Tiled &&       \textbf{94.0} &        \textbf{84.9} &       \textbf{80.6} &        \textbf{74.9} &       \textbf{56.4} &        \textbf{53.0} &&        86.9 &        91.0 &       74.0 &        72.9 &       42.6 &        44.8 \\
        & Sliding &&       84.8 &        86.1 &       70.3 &        72.1 &       44.3 &        47.7 &&        78.3 &        92.7 &       70.3 &        82.5 &       45.4 &        53.1 \\
        & Smoothed &&       91.4 &        74.9 &       70.4 &        60.2 &       39.6 &        36.5 &&        \textbf{90.3} &        \textbf{91.5} &       \textbf{77.8} &        \textbf{76.6} &       \textbf{51.0} &        \textbf{50.8} \\
	    \midrule
        \multirow{3}*{\makecell[l]{In-the-\\wild}}  &  Tiled  &&       90.7 &        81.5 &       \textbf{76.3} &        \textbf{68.3} &       56.7 &        49.7 &&        \textbf{90.8} &        \textbf{84.2} &       \textbf{80.5} &        \textbf{74.4} &       67.9 &        63.5 \\
        & Sliding  &&        82.7 &        80.5 &       70.6 &        64.7 &       \textbf{60.9} &        \textbf{54.7} &&        79.0 &        85.1 &       67.2 &        72.7 &       62.8 &        66.5 \\
        & Smoothed  &&        \textbf{90.8} &        \textbf{72.4} &       73.3 &        56.4 &       53.1 &        41.9 &&       90.0 &        78.7 &       77.0 &        67.5 &       \textbf{72.2} &        \textbf{62.6}  \\
        
        \bottomrule
    \end{tabular}}
    \label{tab:tblsegNew}
\end{table}
}

\newcommand{\tblnnsstats}{
\begin{table}[htb!]
  \caption{Biographical data and NNS and pacifier event statistics for the 10 pacifier-using infants from our NNS clinical in-crib study, six of whom ($\star$) engaged in enough NNS activity for use in machine learning. \textit{Age} is at time of video capture, \textit{BGA} the birth gestational age, \textit{BWt} the birth weight, \textit{Dur} the cumulative event duration, \textit{C-$\kappa$} the Cohen $\kappa$ annotator agreement (incidence on 10 s windows), \textit{Ct} the count, and \textit{Len} the length of individual events. Biographical data are self-reported (hence whole numbers), and event data are averaged from the two annotators (hence fractional counts).}
    \centering
    \ra{1.1} 
    \resizebox{\columnwidth}{!}{\begin{tabular}{@{}lrrrrrrrrrrrrrrrrrrrrr@{}}
        \toprule
        & & \multicolumn{4}{c}{Biographical Data} & & Vid & & \multicolumn{6}{c}{\text{NNS Events (During Pacifier Use)}} & \phantom{a} & \multicolumn{6}{c}{\text{Pacifier Events}}\\ 
        \cmidrule[0.5pt]{3-6} \cmidrule[0.5pt]{8-8} \cmidrule[0.5pt]{10-15} \cmidrule[0.5pt]{17-22}
        Sbj & & Sex & Age & BGA & BWt && Dur && C-$\kappa$ & Ct & Ct/h & Dur & Dur/h & Len & & C-$\kappa$ & Ct & Ct/h & Dur & Dur/h & Len\\
         &&& d & wk & oz && h &&& \# & \#/h & min & min/h & s & &  & \# & \#/h & min & min/h & min\\
        \midrule
        R1$\star$ && M & 100 & 40 & 130 && 10.9 && 0.74 & 636.0 & 58.4 & 39.5 & 3.6 & 3.9 && 0.92 & 13.0 & 1.2 & 219.5 & 20.1 & 16.9\\
        R3$\star$ && F & 98  & 39 & 106 && 11.3 && 0.69 & 490.5 & 43.4 & 51.2 & 4.5 & 6.3 && 0.98 & 10.0 & 0.9 & 270.0 & 23.9 & 27.3\\
        R7$\star$ && F & 103 & 39 & 109 && 13.5 && 0.91 & 817.0 & 60.4 & 60.7 & 4.5 & 4.4 && 1.00 & 5.0 & 0.4 & 214.1 & 15.8 & 44.6\\
        R9         && M & 82  & 40 & 145 && 3.1 && 0.83 & 18.0 & 5.8 & 1.8 & 0.6 & 6.4 && 1.00 & 9.0 & 2.9 & 4.5 & 1.4 & 0.5\\
        R10$\star$ && F & 114 & 39 & 121 && 5.0 && 0.90 & 92.5 & 18.6 & 5.2 & 1.1 & 3.6 && 1.00 & 7.0 & 1.4 & 28.7 & 5.8 & 4.1\\
        R12        && F & 112 & 41 & 101 && 13.1 && 0.96 & 79.5 & 6.1 & 6.3 & 0.5 & 4.8 && 0.99 & 3.5 & 0.3 & 14.3 & 1.1 & 4.2\\
        R15        && F & 102 & 40 & 110 && 14.7 && 0.86 & 106.0 & 7.2 & 14.1 & 1.0 & 8.1 && 0.99 & 7.0 & 0.5 & 52.1 & 3.6 & 8.1\\
        R18$\star$ && F & 142 & 37 & 99  && 6.3 && 0.84 & 115.0 & 18.2 & 6.4 & 1.0 & 3.3 && 0.99 & 4.0 & 0.6 & 39.3 & 6.2 & 10.5\\
        R23        && F & 151 & 42 & 106 && 12.8 && 0.79 & 30.0 & 2.3 & 1.6 & 0.1 & 3.2 && 0.94 & 7.5 & 0.6 & 8.9 & 0.7 & 1.2\\
        R24$\star$ && M & 120 & 39 & 129 && 10.7 && 0.80 & 527.5 & 49.4 & 67.2 & 6.3 & 8.1 && 0.97 & 6.5 & 0.6 & 232.3 & 21.7 & 35.9\\
        \midrule
        Mean       && -  & 112.4 & 39.8 & 115.6 && 10.1 && 0.83 & 291.2 & 27.0 & 25.4 & 2.3 & 5.2 && 0.98 & 7.2 & 0.9 & 108.4 & 10.0 & 15.3 \\
        Std       && -  & 20.8  & 1.4  & 15.0  && 3.9 && 0.08 & 295.0 & 23.4 & 26.3 & 2.2 & 1.9 && 0.03 & 2.9 & 0.8 & 110.0 & 9.3 & 15.5\\
        \midrule
        Mean$\star$ && - & 112.8 & 38.9 & 115.7 && 9.6 && 0.81 & 446.4 & 41.4 & 38.4 & 3.5 & 4.9 && 0.98 & 7.6 & 0.8 & 167.3 & 15.6 & 23.3 \\
        Std$\star$ && - & 16.4  & 1.0  & 12.9  && 3.3 && 0.09 & 288.8 & 18.9 & 26.9 & 2.1 & 1.9 && 0.03 & 3.4 & 0.4 & 105.1 & 7.9 & 15.5\\
        \bottomrule
    \end{tabular}}
    \label{tab:nns-stats}
\end{table}
}

}

\newcommand{\nnsSignal}{
\begin{figure}[t]
\centering
\includegraphics[width=.9\linewidth]{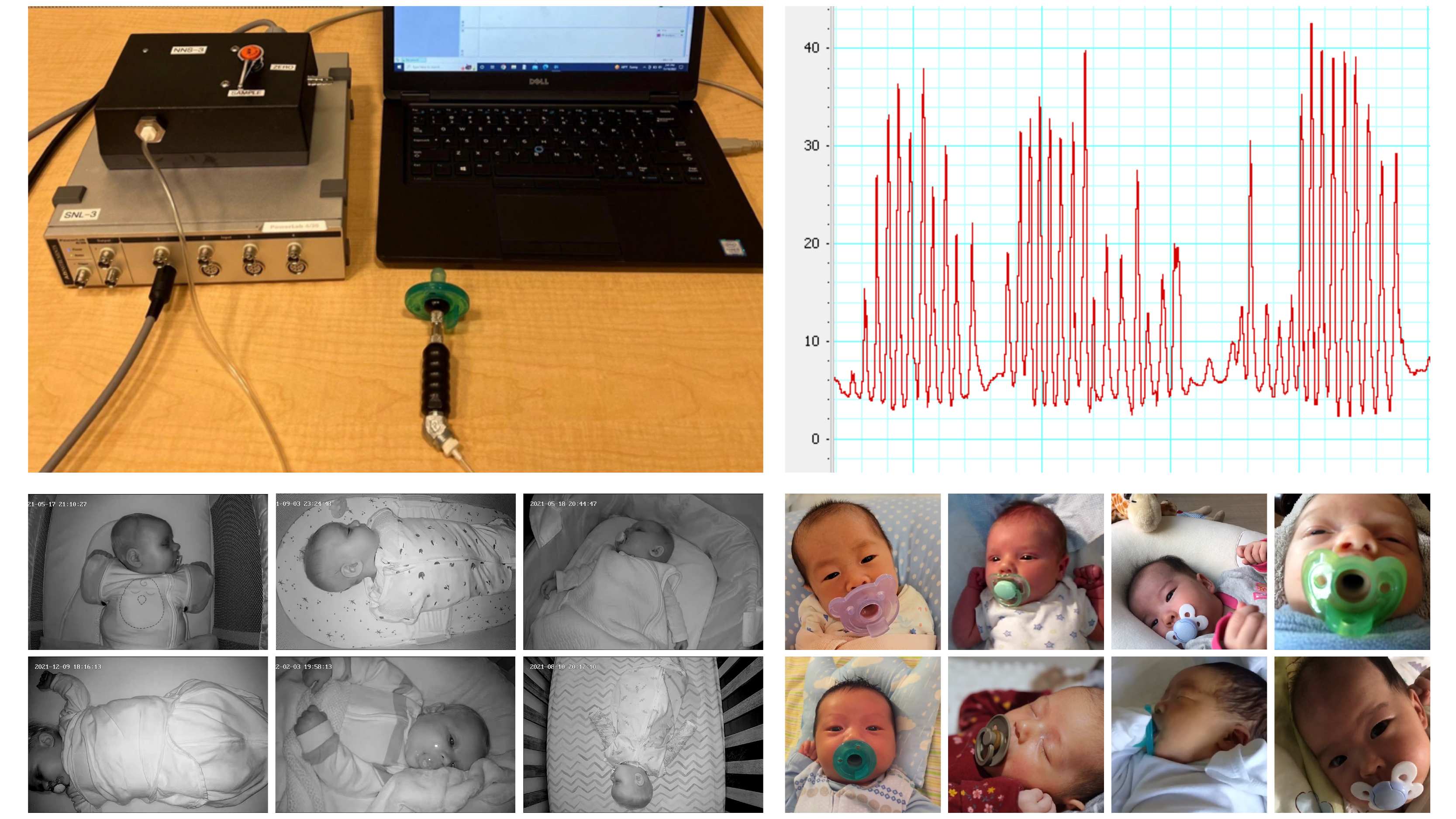}
\caption{\textit{Top:} Illustration of non-nutritive sucking (NNS) signal extracted from a pressure transducer pacifier device \cite{martens_changes_2020}. Our computer vision-based NNS recognition and segmentation algorithms enable algorithmic identification of the relatively rare periods of high NNS activity from long videos, facilitating subsequent clinical expert evaluation. \textit{Bottom:} Still frames from our NNS clinical in-crib dataset \textit{(left)} and our public NNS in-the-wild dataset \textit{(right)}.}
\label{fig:nnsSignal}
\end{figure}
}

\newcommand{\figDiagram}{
\begin{figure*}[t]
\centering
\includegraphics[width=.9\linewidth]{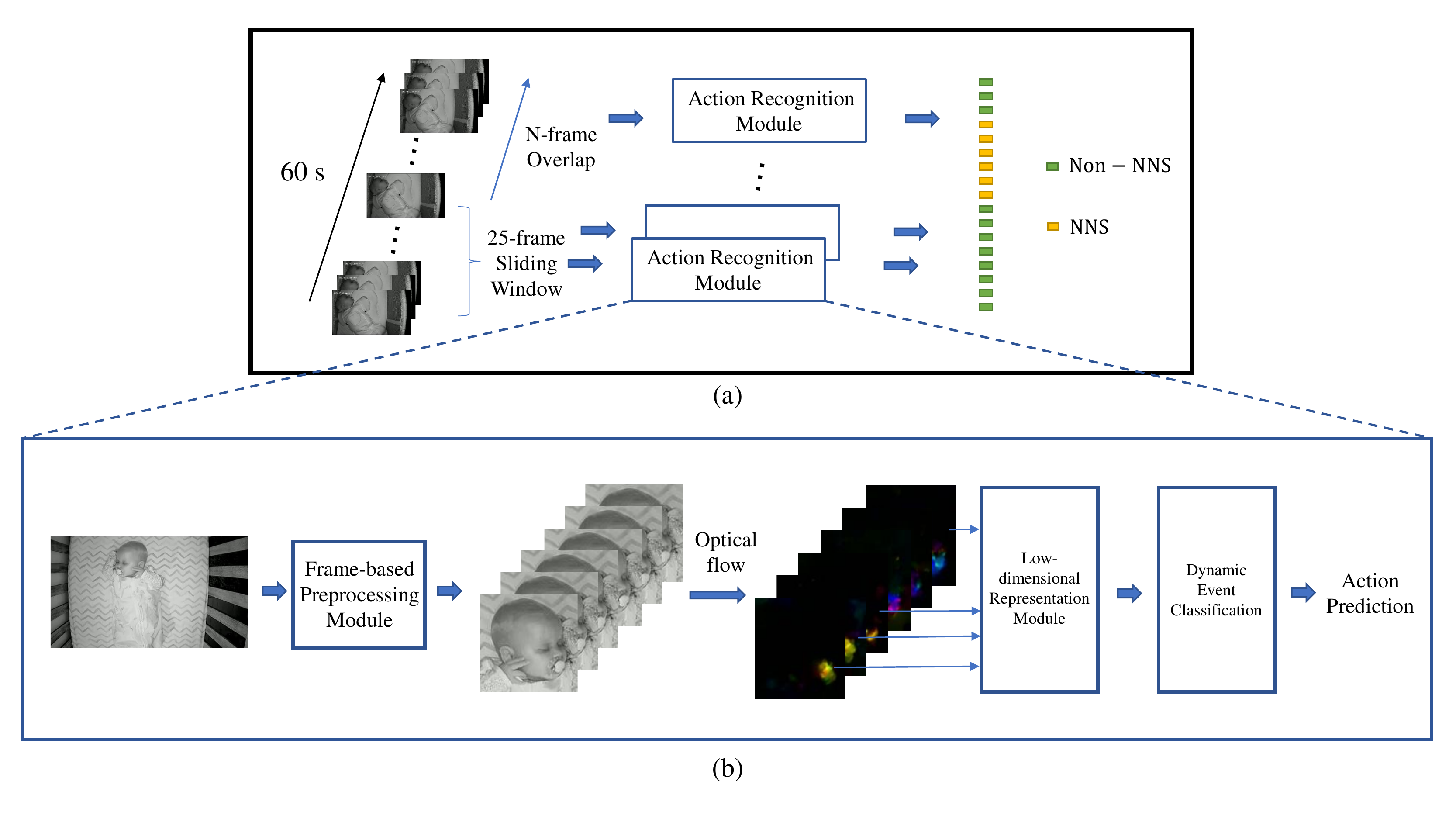}
\caption{
\textit{(a):} Proposed NNS segmentation pipeline, based on aggregating local results of NNS action recognition in sliding windows.
\textit{(b):} Proposed NNS action recognition pipeline, which applies dense optical flow to preprocessed frames, and passes features through a convolutional layer followed by a temporal layer to obtain an action prediction based on spatiotemporal information.}
\label{fig:Diagram}
\end{figure*}
}

\newcommand{\segVis}{
\begin{figure*}[h]
\centering
\includegraphics[width=\linewidth]{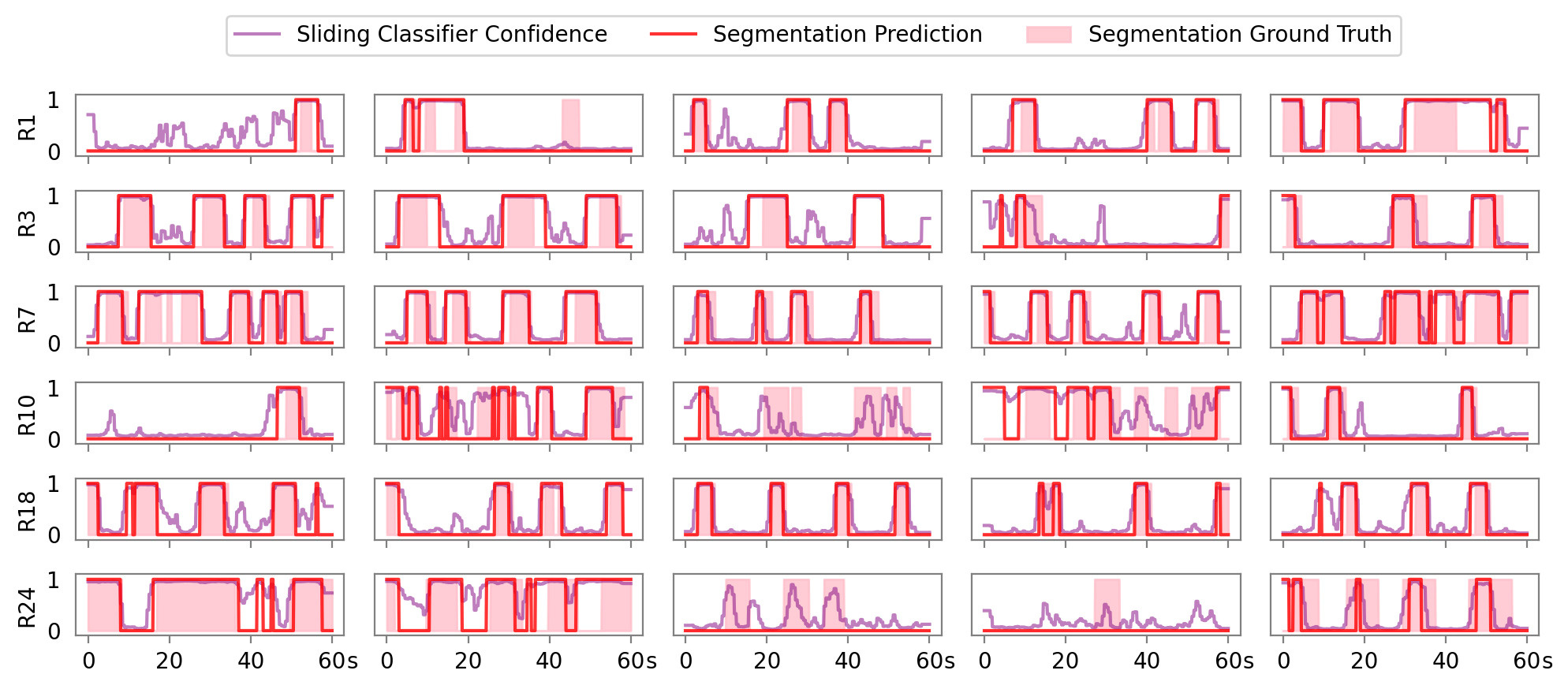}

\caption{Segmentation predictions and ground truth for each 60 s mixed clip from the NNS clinical in-bed dataset, under the sliding window aggregation model configuration and with a confidence threshold of 0.9, boosting precision at the cost of recall.}
\label{fig:segmentation-visualization}
\end{figure*}
}

\newcommand{\setup}{
\begin{figure*}[htb!]
\centering
\includegraphics[width=0.6\linewidth]{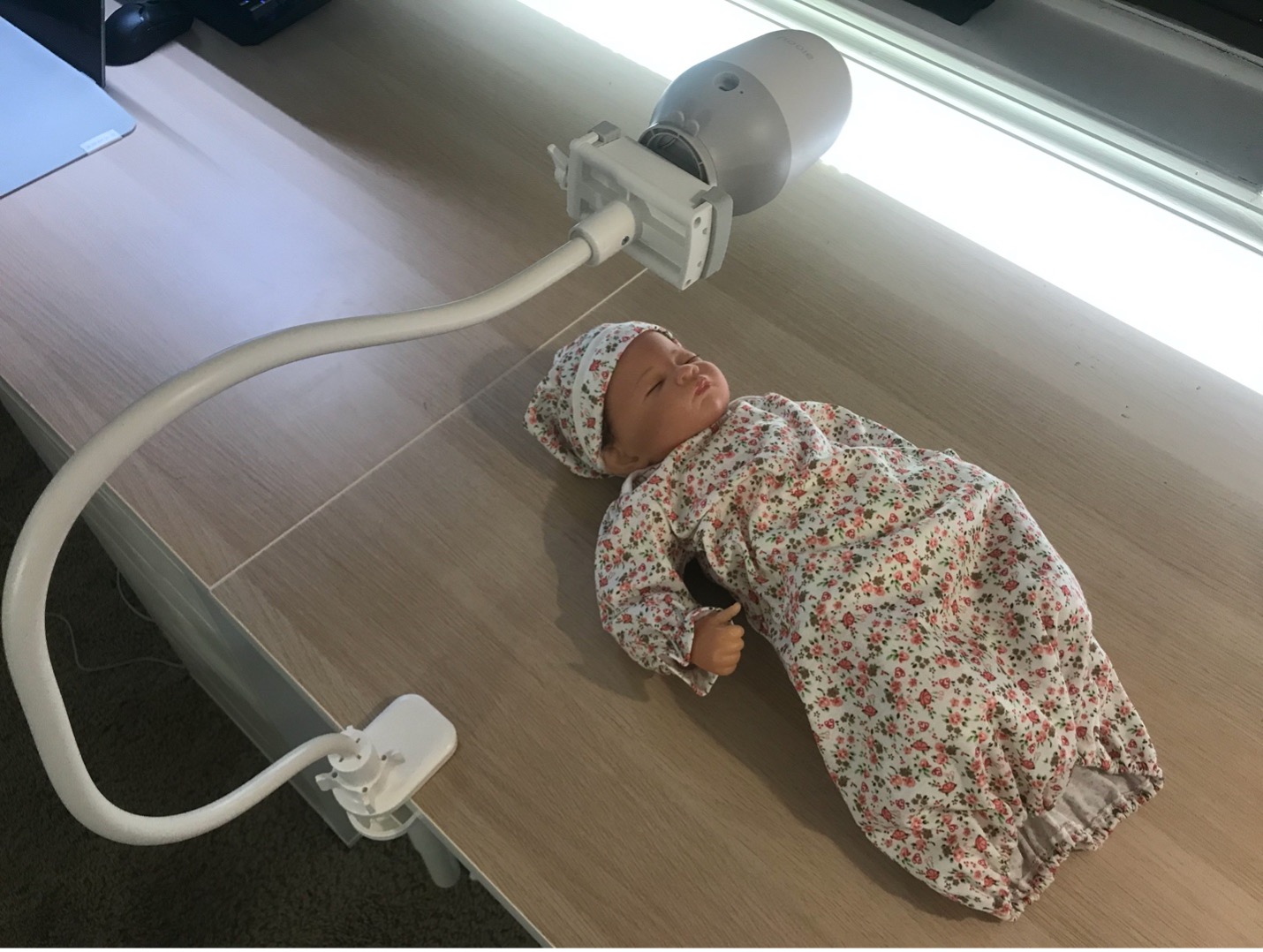}
\caption{Suggested baby monitor placement for study participants for our NNS clinical in-crib dataset. Videos were shot by parents or caregivers in 2021 and 2022 during the pandemic. They are long and feature a wide variety of natural infant behavior, including napping, sleeping, tossing and turning, crying, and caregiver interactions such as pacifier insertion, patting, removal from crib, and more, yielding a true-to-life but technically challenging data source.}
\label{fig:setup}
\end{figure*}
}

\newcommand{\annotSoft}{
\begin{figure*}[htb!]
\centering
\includegraphics[width=0.8\linewidth]{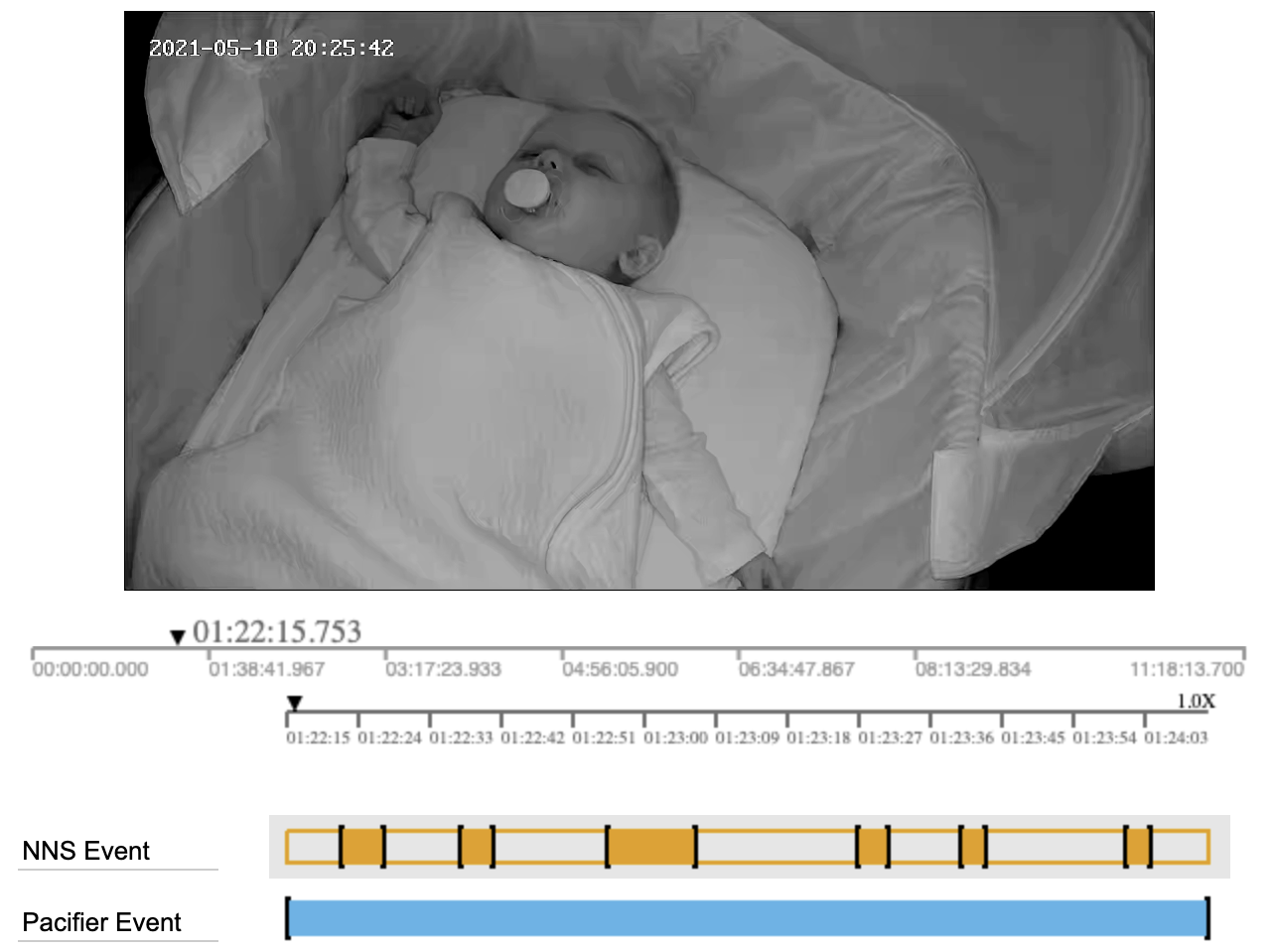}
\caption{NNS annotation tool \cite{dutta2019vgg} used by our behavioral coding team specifically trained for this task. For the NNS clinical in-crib dataset annotations, duplicate coding and systematic checks were implemented to ensure reliability over the hundreds of hours of footage.}
\label{fig:annotSoft}
\end{figure*}
}

\newcommand{\opFlow}{
\begin{figure*}[htb!]
\centering
\includegraphics[width=\linewidth]{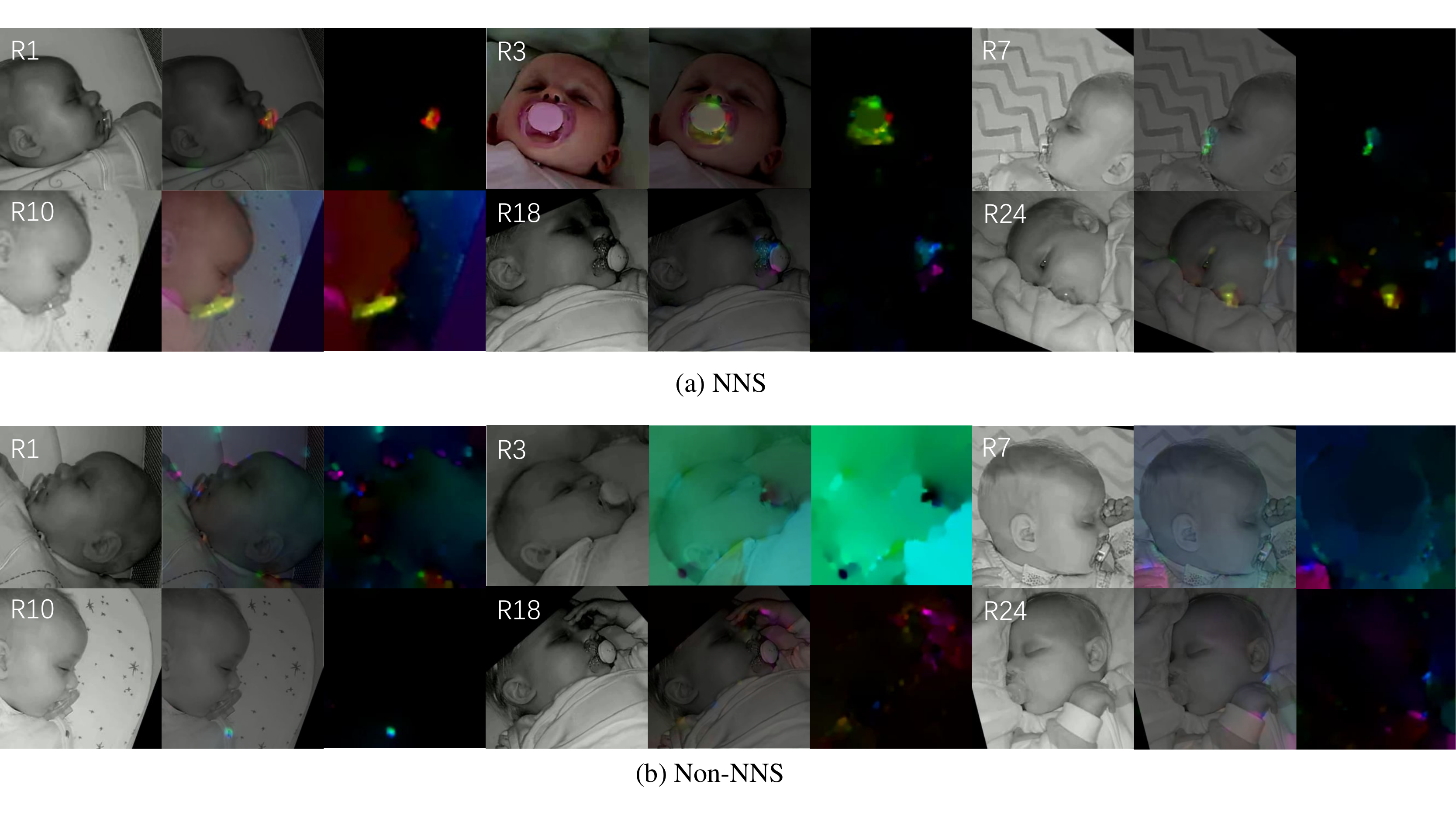}
\caption{Examples of optical flow processing on NNS clinical in-crib data, highlighting its effectiveness at distinguishing (a) NNS events from (b) non-NNS events. See the optical flow ablation study in \secref{nnsResults} for corresponding numerical results.}
\label{fig:opFlow}
\end{figure*}
}

\newcommand{\supp}{
\newpage
\onecolumn
\newcommand{\beginsupplement}{%
        \setcounter{table}{0}
        \setcounter{equation}{0}
        \renewcommand{\theequation}{S\arabic{equation}}
        \setcounter{section}{0}
        \renewcommand{\thetable}{S\arabic{table}}%
        \setcounter{figure}{0}
        \renewcommand{\thefigure}{S\arabic{figure}}%
}
\newcommand{\independent}{\protect\mathpalette{\protect\independenT}{\perp}}
\def\independenT##1##2{\mathrel{\rlap{$##1##2$}\mkern2mu{##1##2}}}
\renewcommand\thesection{\Alph{section}}
\beginsupplement
\newcommand{\MYhref}[3][blue]{\href{##2}{\color{##1}{##3}}}

\section{Supplementary Material} 

\setup
\annotSoft
\tblnnsstats
\opFlow

}

\newcommand{\figref}[1]{Fig.~\ref{fig:#1}}
\newcommand{\tabref}[1]{Table~\ref{tab:#1}}
\newcommand{\secref}[1]{Section~\ref{sec:#1}}

\newcommand{\ra}[1]{\renewcommand{\arraystretch}{#1}} 

\begin{document}
\title{A Video-based End-to-end Pipeline\\for Non-nutritive Sucking Action  Recognition\\and Segmentation in Young Infants}
%
\titlerunning{Non-nutritive Sucking Action Recognition}
%

\author{Shaotong Zhu\inst{1}, Michael Wan\inst{1,2}, Elaheh Hatamimajoumerd\inst{1}, Kashish Jain\inst{1}, Samuel Zlota\inst{1}, Cholpady Vikram Kamath\inst{1}, Cassandra B. Rowan\inst{5}, Emma C. Grace\inst{5}, Matthew S. Goodwin\inst{3,4}, Marie J. Hayes\inst{5}, Rebecca A. Schwartz-Mette\inst{5}, Emily Zimmerman\inst{4}, Sarah Ostadabbas\inst{1}$^*$}

\authorrunning{Shaotong Zhu et al.}

\institute{Augmented Cognition Lab, Department of Electrical \& Computer Engineering\\Northeastern University, Boston MA, USA\\
\and
Roux Institute, Northeastern University, Portland ME, USA\\
\and
Khoury College of Computer Sciences, Northeastern University, Boston MA, USA\\
\and
Bouv\'e College of Health Sciences, Northeastern University, Boston MA, USA\\
\and
Psychology Department, University of Maine, Orono ME, USA\\
$^*$Corresponding author: \email{ostadabbas@ece.neu.edu}
}

\maketitle              
\begin{abstract}
We present an end-to-end computer vision pipeline to detect non-nutritive sucking (NNS)---an infant sucking pattern with no nutrition delivered---as a potential biomarker for developmental delays, using off-the-shelf baby monitor video footage. One barrier to clinical (or algorithmic) assessment of NNS stems from its sparsity, requiring experts to wade through hours of footage to find minutes of relevant activity. Our NNS activity segmentation algorithm solves this problem by identifying periods of NNS with high certainty---up to 94.0\% average precision and 84.9\% average recall across 30 heterogeneous 60 s clips, drawn from our manually annotated NNS clinical in-crib dataset of 183 hours of overnight baby monitor footage from 19 infants. Our method is based on an underlying NNS action recognition algorithm, which uses spatiotemporal deep learning networks and infant-specific pose estimation, achieving 94.9\% accuracy in binary classification of 960 2.5 s balanced NNS vs. non-NNS clips. Tested on our second, independent, and public NNS in-the-wild dataset, NNS recognition classification reaches 92.3\% accuracy, and NNS segmentation achieves 90.8\% precision and 84.2\% recall\footnote{Our code and the manually annotated NNS in-the-wild dataset can be found at \url{https://github.com/ostadabbas/NNS-Detection-and-Segmentation}. Supported by MathWorks and NSF-CAREER Grant \#2143882.}.

\end{abstract}

\section{Introduction}
\label{sec:intro}
Non-nutritive sucking (NNS) is an infant oral sucking pattern characterized by the absence of nutrient delivery \cite{ctx16694780230001401}. NNS reflects neural and motor development in early life \cite{medoff-cooper_neonatal_1995} and may reduce the risk of SIDS \cite{psaila_infant_2017,zavala_abed_how_2020}, the leading cause of death for US infants aged 1-12 months \cite{carlin2017risk}. However, studying the relationship between NNS patterns and breathing, feeding, and arousal during sleep has been challenging due to the difficulty of measuring the NNS signal.

\nnsSignal
NNS occurs in bursts of 6--12 sucks at 2 Hz per suck, with bursts happening a few times per minute during high activity periods \cite{zimmerman_changes_2020}. However, active periods are sporadic, representing only a few minutes per hour, creating a burden for researchers studying characteristics of NNS. Current transducer-based approaches (see \figref{nnsSignal}) are effective, but expensive, limited to research use, and may affect the sucking behavior \cite{zimmerman_patterned_2017}. This motivates our development of an end-to-end computer vision system to recognize and segment NNS actions from lengthy videos, enabling applications in automatic screening and telehealth, with a focus on high precision to enable periods of sucking activity to be reliably extracted for analysis by human experts. 

Our contributions tackle the fine-grained NNS action recognition problem of classifying 2.5 s video clips, and the NNS action segmentation problem of detecting NNS activity in minute-long video clips. The action recognition method uses convolutional long short-term memory networks for spatiotemporal learning. We address data scarcity and reliability issues in real-world baby monitor footage using tailored infant pose state estimation, focusing on the face and pacifier region, and enhancing it with dense optical flow. The action segmentation method aggregates local NNS recognition signals from sliding windows.

We present two new datasets in our work: the \textbf{NNS clinical in-crib dataset}, consisting of 183 h of nighttime in-crib baby monitor footage collected from 19 infants and annotated for NNS activity and pacifier use by our interdisciplinary team of behavioral psychology and machine learning researchers, and the \textbf{NNS in-the-wild dataset}, consisting of 10 naturalistic infant video clips annotated for NNS activity. \figref{nnsSignal} displays sample frames from both datasets.
Our main contributions are (1) creation of the first infant video datasets manually annotated with NNS activity; (2) development of an NNS classification system using a convolutional long short-term memory network, aided by infant domain-specific face localization, video stabilization, and customized signal enhancement, and (3) successful NNS segmentation on longer clips by aggregating local NNS recognition results across sliding windows.
\section{Related Work}
Current methods for measuring the NNS signal are limited to the pressured transducer approach \cite{zimmerman_changes_2020}, and a video-based approach that uses facial landmark detection to extract the jaw movement signal \cite{huang2019infant}. The latter relies on a 3D morphable face model \cite{huber2016multiresolution} learned from adult face data, limiting its accuracy, given the domain gap between infant and adult faces \cite{wan_infanface_2022}; its output also does not directly address NNS classification or segmentation. Our approach offers an efficient, end-to-end solution for both tasks and is freely available.

Action \textit{recognition} is the task of identifying the action label of a short video clip from a set of predetermined classes. In our case, we wish to classify short infant clips based on the presence or absence of NNS. As with many action recognition algorithms, our core model is based on extending 2D convolutional neural networks to the temporal dimension for spatiotemporal data processing.
In particular, we make use of sequential networks (such as long short-term memory (LSTM) networks) after frame-wise convolution to enhance medium-range temporal dependencies \cite{yue2015beyond}.
Action \textit{segmentation} is the task of identifying the periods of time during which specified events occur, often from longer untrimmed videos containing mixed activities. We follow an approach to action segmentation common in limited-data contexts, which patches together signals obtained by a local low-level layer---our NNS action recognition---to obtain a global segmentation result \cite{ding_temporal_2022}.

\section{Infant NNS Action Recognition and Segmentation}
\label{sec:methods}


\figDiagram

\figref{Diagram} illustrates our pipeline for NNS action segmentation, which takes in long-form videos of infants using pacifiers and predicts timestamps for NNS events in the entire video. We first cover the long video with short sliding windows, then apply our NNS action recognition module to obtain a classification for NNS vs non-NNS (or the confidence score), and aggregate the output classes (or scores) into a segmentation result consisting of predicted start and end timestamps. 


\subsection{NNS Action Recognition}
\label{sec:method-classification}

The technical core of our model is the NNS action recognition system, depicted in \figref{Diagram}b, which is itself composed of a frame-based preprocessing module followed by a spatiotemporal classifier. The preprocessing module uses a pre-trained model; only the spatiotemporal classifier is actively trained with our data.

\subsubsection{Preprocessing Module} 
Our frame-based preprocessing module applies the following transformations in sequence. All three steps are used to produce training data for the subsequent spatiotemporal classifier, but during inference, the data augmentation step is not applicable and is omitted.

\textit{Smooth facial crop.} The RetinaFace face detector \cite{retinaface2019} is applied to frames in each clip until a face bounding box is found and propagated to earlier and later frames using the minimum output sum of squared error (MOSSE) tracker \cite{mossepaper}. To smooth the facial bounding box sequence and address temporal discontinuity, saliency corners \cite{shi1994good} are detected from the initial frame and tracked to the next frame using the Lucas-Kanade optical flow algorithm \cite{lucas1981iterative}. The trajectory is smoothed using a moving average filter and applied to each bounding box to stabilize the facial area. The raw input video is then cropped to this smoothed bounding box, resulting in a video featuring the face alone.

\textit{Data augmentation.} When preprocessing videos to create training data for the spatiotemporal classifier, we apply random transformations such as rotations, scaling, and flipping to the face-cropped video, to improve generalizability in our data-limited setting.  


\textit{Optical flow.} 
After trimming and augmentation, we calculate the short-time dense optical flow \cite{liu2009beyond} between adjacent frames, and map the results into the hue, saturation, and value (HSV) color space by cascading the optical flow direction vector and magnitude of each pixel. This highlights the apparent motion between frames, magnifying subtle NNS movements (as illustrated in Supp. \figref{opFlow}.). 


\subsubsection{Spatiotemporal-based Action Classifier} 
Finally, the optical flow video is processed by a spatiotemporal model that outputs an action class label (NNS or non-NNS). Two-dimensional convolutional neural networks extract spatial representations from static images, which are then fed in sequence to a temporal convolution network for spatiotemporal processing. The final classification outcome is the output of the last temporal convolution network unit.

\subsection{NNS Action Segmentation}
\label{sec:action-segmentation}
To segment NNS actions in mixed videos with transitions between NNS and non-NNS activity, we applied NNS recognition in 2.5 s sliding windows and aggregated results to predict start and end timestamps. This window length provides fine-grained resolution for segmentation while being long enough (26 frames at a 10 Hz frame rate) for consistent human and machine detection of NNS behavior. To address concerns about the coarseness of this resolution, we tested the following window-aggregation configurations, the latter two of which have finer 0.5 s effective resolutions:
\begin{description}
    \item \textbf{Tiled:} 2.5 s windows precisely tile the length of the video with no overlaps, and the classification outcome for each window is taken directly to be the segmentation outcome for that window.
    \item \textbf{Sliding:} 2.5 s windows are slid across with 0.5 s overlaps, and the classification outcome for each window is assigned to its (unique) middle-fifth 0.5 s segment as the segmentation outcome.
    \item \textbf{Smoothed:} 2.5 s windows are slid across with 0.5 s overlaps, the classification \textit{confidence score} for each window is assigned to its middle-fifth 0.5 s segment, a 2.5 s moving average of these confidence scores are taken, then the averaged confidence scores are thresholded for the final segmentation outcome.
\end{description}

\section{Experiments, Results, and Ablation Study}
\label{sec:results}

\subsection{NNS Dataset Creation}
\label{sec:data-preparation}

Our primary dataset is the \textbf{NNS clinical in-crib dataset}, consisting of 183 h of baby monitor footage collected from 19 infants during overnight sleep sessions by our clinical neurodevelopment team, with Institutional Review Board (IRB \#17-08-19) approval. Videos were shot in-crib with the baby monitors set up by caregivers, under low-light triggering the monochromatic infrared mode. Tens of thousands of timestamps for NNS and pacifier activity were placed, by two trained behavioral coders per video. For NNS, the definition of an event segment was taken to be an NNS \textit{burst}: a sequence of sucks with ${<}1$ s gaps between. We restrict our subsequent study to NNS during pacifier use, which was annotated more consistently. Cohen $\kappa$ annotator agreement of NNS events during pacifier use (among 10 pacifier-using infants) averaged 0.83 in 10 s incidence windows, indicating strong agreement by behavioral coding standards, but we performed further manual selection to increase precision for machine learning use, as detailed below\footnote{See the Supp. \figref{setup} and Supp. \figref{annotSoft} for more on the creation of the NNS clinical in-crib dataset, and Supp. \tabref{nns-stats} for full Cohen $\kappa$ scores, biographical data, and NNS and pacifier event statistics.}. We also created a smaller but publicly available \textbf{NNS in-the-wild dataset} of 14 YouTube videos featuring infants in natural conditions, with lengths ranging from 1 to 30 minutes, and similar annotations.

From each of these two datasets, we extracted 2.5 s clips for the classification task and 60 s clips for the segmentation task. In the NNS clinical in-crib dataset, we restricted our attention to six infant videos containing enough NNS activity during pacifier use for meaningful clip extraction. From each of these, we randomly drew up to 80 2.5 s clips consisting entirely of NNS activity and 80 2.5 s clips containing non-NNS activity for classification, for a total of 1,600; and five 60 s clips featuring transitions between NNS and non-NNS activity for segmentation, for a total of 30; redrawing if available when annotations were not sufficiently accurate. In the NNS in-the-wild dataset, we restricted to five infants exhibiting sufficient NNS activity during pacifier use, from which we drew 38 2.5 s clips each of NNS and no NNS activity for classification, for a total of 76; and from two to 26 60 s clips of mixed activity from each infant for segmentation, for a total of 39; again redrawing in cases of poor annotations.




\subsection{NNS Recognition Implementation and Results}
\label{sec:nnsResults}
For the spatiotemporal core of our NNS action recognition, we experimented with four configurations of 2D convolutional networks, a 1-layer CNN, ResNet18, ResNet50, and ResNet101 \cite{he2016deep}; and three configurations of sequential networks, an LSTM, a bi-directional LSTM, and a transformer model \cite{vaswani2017attention}. The models were trained for 20 epochs under a learning rate of 0.0001 using PyTorch 1.8.1 with CUDA 10.2, and the best model chosen based on a held-out validation set.

We trained and tested this method with NNS clinical in-crib data from six infant subjects under a subject-wise leave-one-out cross-validation paradigm. Action recognition accuracies under are reported on the top left of \tabref{resultCombo}. The ResNet18-LSTM configuration performed best, achieving $94.9\%$ average accuracy over six infants using optical flow input. The strong performance (${\geq}85.2\%$) across all configurations indicates the viability of the overall method.
We also evaluated a model trained on all six infants from the clinical in-crib dataset on the independent in-the-wild dataset. Results on the bottom left of \tabref{resultCombo} again show strong cross-configuration performance (${\geq}79.5\%$), with ResNet101-Transformer reaching $92.3\%$, demonstrating strong generalizability of the method.

\tblresultCombo


As expected, models trained on the clinical in-crib data test worse on the independent in-the-wild data. But interestingly, models with the smaller ResNet18 network suffered steep drop-offs in performance when tested on the in-the-wild data, while models based on the complex ResNet101 fared better under the domain shift. Beyond this, it is hard to identify clear trends between configurations or capacities and performance.

\subsubsection{Optical Flow Ablation}
Performance of all models with raw RGB input replacing optical flow frames can be found on the right side of \tabref{resultCombo}. The results are weak and close to random guessing, demonstrating the critical role played by optical flow in detecting the subtle NNS signal. This can also be seen clearly in the sample optical flow frames visualized in Supp. \figref{opFlow}.


\subsection{NNS Segmentation Results}

Adopting the best ResNet18-LSTM recognition model, we tested the three configurations of the derived segmentation method on the 60 s mixed activity clips, under the same leave-one-out cross-validation paradigm on the six infants. In addition to the default classifier threshold of 0.5 used by our recognition model, we tested a 0.9 threshold to coax higher precision, as motivated in \secref{intro}. We use the standard evaluation metrics of average precision $\text{AP}_t$ and average recall $\text{AR}_t$ based on hits and misses defined by an intersection-over-union (IoU) with threshold $t$, across common thresholds $t\in\left\{0.1, 0.3, 0.5\right\}$\footnote{We follow the definitions from \cite{idrees_thumos_2017},  except that multiple predictions of a single ground truth event are resolved by the highest IoU, rather than confidence score.}. Averages are taken with subjects given equal weight, and results tabulated in \tabref{tblsegNew}.

The metrics reveal strong performance from all methods and both confidence thresholds on both test sets. Generally, as expected, setting a higher confidence threshold or employing the more tempered tiled or smoothed aggregation methods favours precision, while lowering the confidence threshold or employing the more responsive sliding aggregation method favours recall. The results are excellent at the IoU threshold of 0.1 but degrade as the threshold is raised, suggesting that while these methods can readily perceive NNS behavior, they are still limited by the underlying ground truth annotator accuracy. The consistency of the performance of the model across both cross-validation testing in the clinical in-crib dataset and the independent testing on the NNS in-the-wild dataset suggests strong generalizability. \figref{segmentation-visualization} visualizes predictions (and underlying confidence scores) of the sliding model configuration with a confidence threshold of 0.9, highlighting the excellent precision characteristics and illustrating the overall challenges of the detection problem.


\tblsegNew
\segVis

\section{Conclusion}
We present our novel computer vision method for detection of non-nutritive sucking from videos, with a spatiotemporal action recognition model for classifying short video clips and an action segmentation model for determining event timestamps within longer videos. Our work is grounded in our methodological collection and annotation of infant video data, from both in-crib and in-the-wild videos. We use domain-specific techniques such as dense optical flow and infant state tracking to detect subtle sucking movements and ameliorate a relative scarcity of data. Our successful segmentation results demonstrate the potential for use in research and clinical applications. Future work could improve upon segmentation accuracy or extend our approach to assess other NNS signal characteristics of interest to neurodevelopmental researchers, including individual suck frequency, strength, duration, and general temporal pattern.


\bibliographystyle{splncs04}
\bibliography{mybib}

\supp
\end{document}